\documentclass[conference]{IEEEtran}
\IEEEoverridecommandlockouts
\usepackage{cite}
\usepackage{amsmath,amssymb,amsfonts}
\usepackage{algorithmic}
\usepackage{graphicx}
\usepackage{textcomp}
\graphicspath{{images/}} 
\def\BibTeX{{\rm B\kern-.05em{\sc i\kern-.025em b}\kern-.08em
    T\kern-.1667em\lower.7ex\hbox{E}\kern-.125emX}}
\begin{document}

\title{Potentials and Limitations of Deep Neural
Networks for Cognitive Robots}

\author{Doreen Jirak and Stefan Wermter  \\ Knowledge Technology, Department of Informatics, University of Hamburg, Germany}

\maketitle

\begin{abstract}
Although Deep Neural Networks reached remark-
able performance on several benchmarks and even gained scien-
tific publicity, they are not able to address the concept of cognition
as a whole. In this paper, we argue that those architectures
are potentially interesting for cognitive robots regarding their
perceptual representation power for audio and vision data.
We identify crucial settings for cognitive robotics where deep
neural networks have as yet only contributed little compared
to the challenges in this area. Finally, we highlight the rather
unexplored area of Reservoir Computing for sequence learning.
This new paradigm of learning recurrent neural networks in a
fast and robust way qualifies to be an integral part of cognitive
robots and may inspire novel developments.
\end{abstract} 

\section{Introduction}
With the advent of both data availability and accelerated
GPU computations, deep neural network architectures (DNN)
have flourished over the past years and showed tremen-
dous success on several benchmarks for computer vision,
and audio and speech. However, DNN models demand high
training times for sensitive parameter tuning and network-
specific optimization like regularization techniques, which is
a rather engineering approach. Although DNNs are motivated
by neurobiological principles observed in the visual cortex, the
complete picture of the computations boils down to learning
filters detecting specific frequency characteristics like edges
in images or voice pitch in audio data. As the learning is
carried out in a supervised manner, the use of a huge amount
of labeled data has a crucial impact on high accuracy and other
evaluation measures.
The area of cognitive robotics is highly interdisciplinary
and advances in sensor and robot technology allow the in-
tegration of human-like capabilities. The aim of cognitive
skills in robots is motivated by their increased integration in
human's everyday life, whether as a companion in human-
robot interaction (HRI) or as an autonomous system in health
care and industry.
DNNs can only address part of what actually cognitive
systems need and which challenges they face. The focus
on supervised benchmarking shifts research on the multiple,
interconnected facets of cognitive robots to a rather engineered
approach accompanied by constraints on the data and the
environment. Critical investigations on DNNs also reveal that
due to their high dependence on apriori knowledge, they are
easily fooled \cite{Nguye_15} and fail to cope with, e.g., real-world speech
data from children. Recent advances of (deep) unsupervised
networks closely related to neural information principles like
sparse autoencoders or self organizing maps may stimulate
again joining state-of-the-art algorithms successfully applied
to robot scenarios with deep architectures.
In the present paper, we will discuss the pros and cons
of DNNs in cognitive robotics. We highlight their usefulness
but also limitations, discussing also the potential of Reservoir
Computing in the context of cognitive robots.

\section{Lab Conditions vs. the Real World}

A challenging aspect of cognitive systems and specifically
cognitive robots is the high parallelization of tasks with a vast
amount of incoming stimuli which need to be filtered to trigger
a suitable (re)action. We are able to focus our attention onto a
specific object or sound but we can also be easily distracted as
is observable in the ventriloquist effect, where humans' sound
perception is influenced by a doll being an additional visual
stimulus. Researchers thus developed algorithms to capture
the most salient cues in the scene but either on datasets \cite{Xu_15}
or under very constrained lab conditions \cite{Barro_17}. Although imple-
mentations using deep networks showed promising results in
those limited cases, DNN will have problems with changing
conditions in real-world environments due to retraining issues
inherent in closed set classification system.
A recent robotic experiment on indoor exploration proposed
a framework based on DNN to exploit the robot capabilities to
learn the unknown terrain in a human-like fashion using the
turtlebot platform \cite{Tai_17}. For the robot vision, a Kinect device
was mounted delivering depth information. Although one may
argue that Kinect sensors are error-prone and question their
reliability in a robust image capturing, the availability of the
authors' code enable other researchers to explore further the
advantages and pitfalls of the approach, identifying further
problems and possible solutions enriching discussions in the
cognitive robotic community for this important task.
Another important factor is that learning in humans is
not subject to one specific task for a specific point in time
and space but is continuous throughout life. Although DNNs
produce stable feature representations which may code for
more abstract concepts similar to neural codes along the brain
hierarchy, yet they are less flexible and do not allow online
adaptation nor any memory mechanisms. A new trend, how-
ever, to avoid retraining is using transfer learning: weights of
a deep network successfully trained on a set of e.g. images are
transfered to solve a similar task. A fruitful combination for
robot applications could be to join those strategies with open
set classification algorithms like the growing-when-required
networks (GWR, \cite{Marsl_02}), which allow insertion and deletion of labels.
The benefit of a GWR is that it provides a computational
method to the question how actually learning is guided in the
absence of any labels. The robust representations emerging
from a deep neural network can thus complement the flex-
ible classification in environments beyond the restricted lab
conditions.
\section{Children vs. Adult Learning}

Developmental psychology is an important research area
to gain an understanding on how we acquire cognitive skills
starting from first imitation tasks to controlled motor acts, from
first babbling to words and sentences to finally reasoning,
consciousness and empathy (theory of mind). In contrast to
deep networks being trained on billions of e.g. images for
object recognition, infants learn rather slow and are provided
only with a very limited excerpt of the world. DNNs tuned
for benchmarking as was carried out during the past years
is very different from developmental learning and the ac-
quisition of cognitive skills. Although their performance on
speech recognition tasks is indubitable, DNN models show
low performance when the speech resources are getting noisy
(e.g. repetitions) or are from children, since their language
differs from adults.
In addition, the way children learn is driven by high intrinsic
motivation \cite{Oudey_07}, curiosity \cite{Gottl_13} and their physical interaction with
the world (embodiment). The latter was shown to be essential
in exploring affordances for the associative learning between
objects and grasp types \cite{Oztop_04}, keeping also into account parental
scaffolding \cite{Ugur_15}. The link between motor activation in the brain
and language acquisition speaks in favor of embodied learning
which demands to ground experiments in real robots (see e.g.
\cite{Maroc_10}) rather than a set of labeled data.
Until now, these important aspects to acquire cognitive skills
are absent in learning systems based on DNNs which may
be explained by their difficulty in interpretation. A literature
research on this topic revealed that only a few trends emerged
towards closing this gap. One approach used a generative
deep neural network based on the time course of learning
object categories in children \cite{Saxe_13}. Essentially, the authors
demonstrated that despite using linear neurons, their model
was able to capture the refinement of semantic labels and with
an additional analysis on the temporal dynamics of gradient-
based learning, that this differentiation emerges naturally in
hierarchical networks.
Another recent paper \cite{Sigau_16} identified a gap between de-
velopmental learning and deep neural networks along the
hierarchical processing of object recognition on the perceptual
level for the acquisition of action knowledge on the cognitive
level where affordances play a crucial role. In the light of
embodied cognition, affordances are grounded in sensorimotor
experiences. The authors \cite{Sigau_16} argued in favor of unsupervised
learning techniques as labels of sensorimotor experiences do
not exist. Though deep networks do not allow for flexible,
online adaptation of e.g. grasping when affordances may
change (think about the change of perspective of a mug
handle), the main argument integrating deep unsupervised
models is that they can serve as an enhancement strategy for
learned sensorimotor experience as in humans during sleep.
In conclusion, although still other developmental learning
mechanisms as the mentioned curiosity are missing and can
not be captured by DNNs, they might be a useful tool for
representational learning, encouraging a novel research area
called "deep developmental learning" \cite{Sigau_16}.
Regarding the interpretability issues in DNN, a deep net-
work in a one-shot word learning scenario was employed and
the results investigated by means of methods from cognitive
psychology, namely the "cognitive bias" which allows children
to eliminate word hypotheses from a vast possible word space
\cite{Ritte_17}. The network evaluation, which used training on the
popular ImageNet database, underpinned the ``shape bias"
hypothesis, i.e. that humans tend to label objects similar in
shape as one entity. Joining DNNs with cognitive psychology
may open the way to further investigate the positive as well as
negative aspects of biases in human learning, complementing
behavioral studies which found the basis for many cognitive
robotic scenarios.

\section{Reservoir Computing for Sequential Learning in Robot Scenarios}

Learning sequential tasks is inevitably necessary for the
development of cognitive systems, may it be navigation,
planning or communication with language and gestures. Deep
neural architectures do not possess an explicit time resolution
to capture time correlation nor timescales inherent in sentences
and actions. A popular technique to overcome the missing
temporal link in, e.g., CNNs is to use 3D convolution kernels
for image stacks representing videos. However, the benefit of
such kernels on the performance compared to the standard 2D
kernel is not yet satisfiable answered \cite{Karpa_14}.
Due to the inherent correlations in sequential tasks, Re-
current Neural Networks (RNN) are a widely used tool as
they provide local network memory and are thus able to
compute different time aspects (e.g. continuous-time RNN).
RNNs in their diverse implementations were successfully used
in cognitive robotic scenarios (see e.g. \cite{Oubba_14}, \cite{Murat_17}). Due to
the error computations propagated through the whole network
using gradient descent, RNN training suffers from vanishing or
exploding gradients, trapping into local minima, and network
bifurcations. To overcome these issues, RNNs need a lot of
retraining and are thus computationally demanding.
Alternatively, the Reservoir Computing (RC) paradigm in-
troduced a simplified training by the conceptual separation of
a high-dimensional reservoir of randomly connected neurons
providing rich input representations which are read out by
simple linear models and thus updating the network weights
becomes obsolete. Popular implementations are the Liquid
State Machines \cite{Maass_02a} (LSM) and Echo State Networks \cite{Jaege_01}
(ESN) the latter demonstrating competitive performance in
predictions tasks for (chaotic) timeseries \cite{Lukov_12}. Despite the
reduced computational effort, only a few applications for
robots were developed until today. Some highlights in the
literature using the RC principles include navigation \cite{Dasgu_13}, \cite{Anton_15}
and language acquisition \cite{Hinau_14}. Especially the latter research
shows that RC is a promising framework combining principles
from neuroscience and developmental psychology for the
highly complex cognitive task of learning and understanding
language. Also, only little is known about the potential of
RC algorithms for visual tasks. As images and videos are
highly complex in structure and introduce varying lighting
conditions and perspectives, a solution combining the robust
feature capabilities provided by DNNs with ESNs capturing
the inherent temporal structure showed good performance on
a gesture recognition task for a set of command gestures \cite{Jirak_15}.
The coupling of DNNs with RC was also highlighted recently
regarding future research directions \cite{Gouda_16}. Methods based on
RC can also be exploited for robot control \cite{Poly_15} and locomotion
\cite{Wyffe_09} but coupling RC with more complex robot scenarios is
still in its infancy.
We believe, that the area of cognitive robotics would sub-
stantially benefit from further research on the potentials of RC
frameworks complementing other computational methods like
DNNs for robot perception or even substituting existing RNN
methods exploiting the simplified yet robust training. Also,
analysis of ESNs regarding their "edge of stability" would
unify hypotheses on cognition underlying chaotic computa-
tions with performance maximization \cite{Bianch_16a}, which can lead to
effective network design.
\section{Conclusion}
The argument in favor of cognitive systems is diametral to
the current benchmarking and competition of accuracies using
deep neural networks. Although those network architectures
are beneficial for perceptual or supervised learning involved
in the acquisition of cognitive skills, we highlighted some
limitations and suggest alternative computations for future ap-
plications. This brief communication focused on recent neural
network approaches and Reservoir Computing to stimulate
discussions on the integration of those methods into cognitive
robots. Our intention was not to exclude other important
methods like probabilistic models or dynamic neural fields. We
rather argue, that advances in technology in humans everyday
life demand tighter coupling in the community regarding
current progress in the neural network and machine learning
area with cognitive models proven to be successful in the
development of cognitive robots.

\bibliographystyle{IEEEtran}
\bibliography{eucog17}
\end{document}